\documentclass[sigconf]{acmart}

\usepackage{subcaption}
\usepackage{booktabs} 
\usepackage[colorinlistoftodos]{todonotes}
\usepackage{multirow}

\begin{document}
\copyrightyear{2018} 
\acmYear{2018} 
\setcopyright{acmcopyright}
\acmConference[FDG'18]{FDG}{August 7-10, 2018}{Malm\"o, Sweden}
\acmBooktitle{Foundations of Digital Games 2018 (FDG18), August 7--10, 2018, Malm\"o, Sweden}
\acmPrice{15.00}
\acmDOI{10.1145/3235765.3235781}
\acmISBN{978-1-4503-6571-0/18/08}
\title{Deep Unsupervised Multi-View Detection of Video Game Stream Highlights}

\author{Charles Ringer}
\affiliation{%
    \department{Department of Computing}
  \institution{Goldsmiths, University of London}
  \city{London}
  \state{United Kingdom}
}
\email{c.ringer@gold.ac.uk}

\author{Mihalis A. Nicolaou}
\affiliation{%
    \department{Department of Computing}
  \institution{Goldsmiths, University of London}
  \city{London}
  \state{United Kingdom}
}
\email{m.nicolaou@gold.ac.uk}
\renewcommand{\shortauthors}{Ringer and Nicolaou}

\begin{abstract}
We consider the problem of automatic highlight-detection in video game streams.  Currently, the vast majority of highlight-detection systems for games are triggered by the occurrence of hard-coded game events (e.g., score change, end-game), while  most advanced tools and techniques are based on detection of highlights via visual analysis of game footage.  We argue that in the context of game streaming, events that may constitute highlights are not only dependent on game footage, but also on social signals that are conveyed by the streamer during the play session (e.g., when interacting with viewers, or when commenting and reacting to the game).  In this light, we present a multi-view unsupervised deep learning methodology for novelty-based highlight detection.  The method jointly analyses both game footage and social signals such as the players facial expressions and speech, and shows promising results for generating highlights on streams of popular games such as \textit{Player Unknown's Battlegrounds}. 
\end{abstract}

%
%

\ccsdesc[500]{Computing methodologies~Scene anomaly detection}
\ccsdesc[500]{Computing methodologies~Neural networks}
\ccsdesc[300]{Computing methodologies~Scene understanding}
\ccsdesc[300]{Computing methodologies~Video segmentation}

\keywords{Video game stream analysis, highlight detection, event detection}

\maketitle

\section{Introduction}
Recently, live  streaming services such as \texttt{TWITCH.TV}\footnote{www.twitch.tv}, \texttt{Youtube Gaming}\footnote{gaming.youtube.com}, and \texttt{Huya}\footnote{www.huya.com} have become popular platforms for video game players to broadcast themselves playing on the Internet.  During a typical stream, players broadcast both game footage, as well as video of their face via a web-cam, while also communicating with viewers via audio and text chat.


In this work, we present the first, unsupervised, multi-modal, approach towards generating highlight clips, by analyzing both audio and video arising from the player's camera feed, as well as game footage (both video and audio), in order to identify novel events occurring during a stream. We use convolutional autoencoders for visual analysis of game scene and face, spectral features and component analysis for audio, while recurrent layers are utilized for fusing representations and eventually, detecting highlights on multi-view time-series data.

\begin{figure}[t]
\includegraphics[width=0.8\columnwidth]{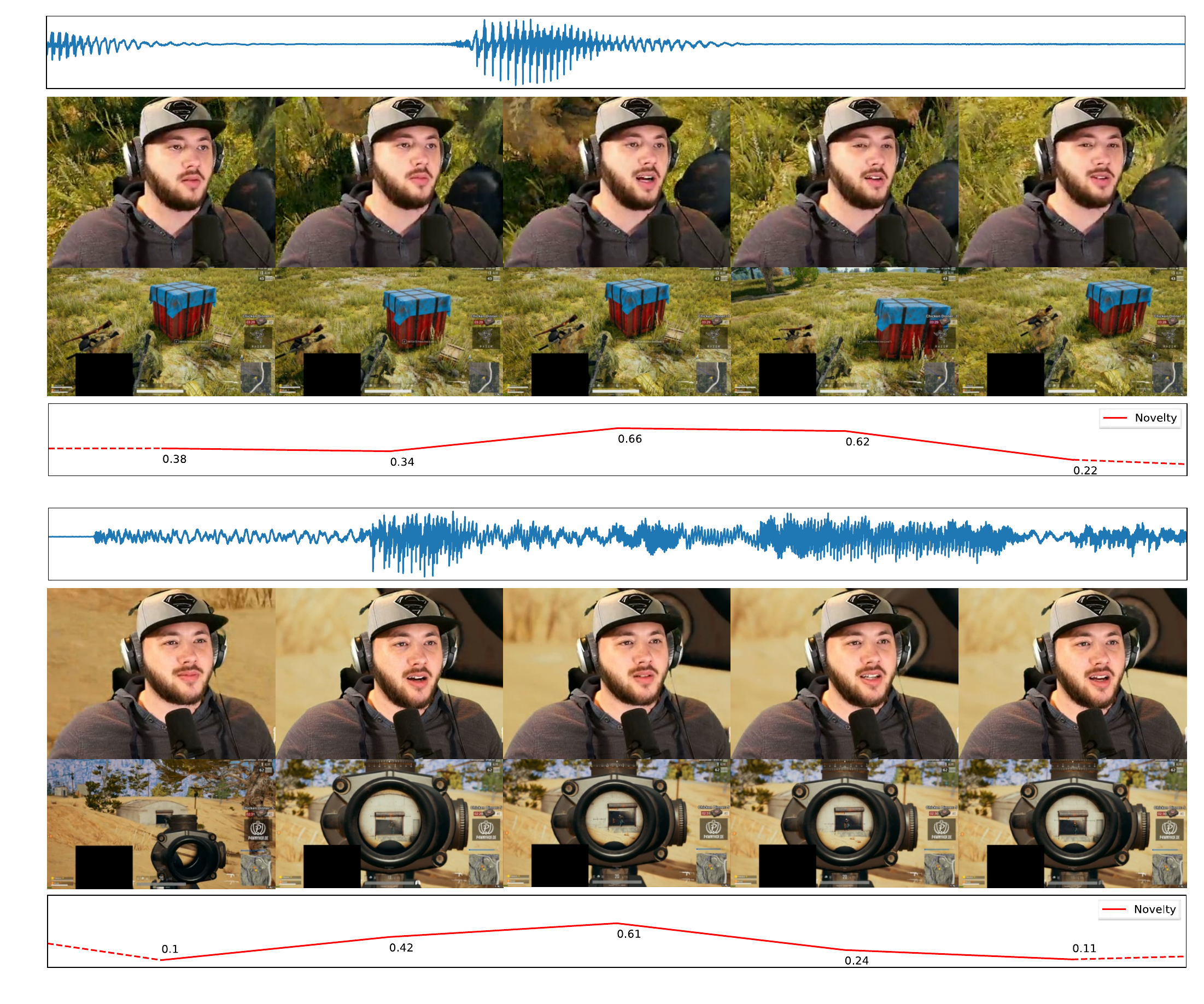}
\caption{Example highlights. Audio waveform, face frames, game frames, and prediction error shown. Above: The streamer makes a joke after picking up a good item at a crate. Below: The streamer initiates a fire fight with another player. In both cases the highlight apex is the center frame.
}
\label{fig:intro_fig}
\end{figure}

\begin{figure*}[t]
\centering
\includegraphics[width=0.9\textwidth]{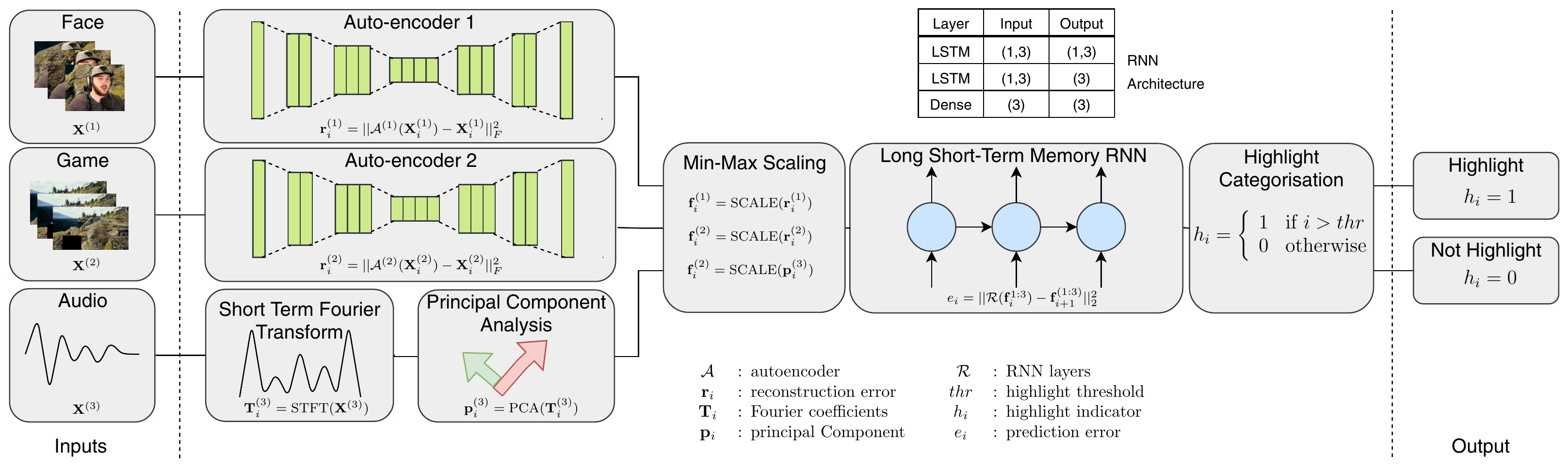}
\caption{Overview of the system}
\label{fig:sysoverview}
\end{figure*}

\section{Related Work}
\subsection{Event and Highlight Detection}
Detecting events in audio-visual data is an active area of research across a range of research domains. Perhaps the most pertinent to this study is Chu et al. \cite{Chu2015, Chu2017} who, studying \textit{League of Legends} tournament streams, used in-game messages to select events and various motion based features, such as monitoring particle effects, to detect highlights.

Much event detection research has been focused on motion. Simonyan and Zisserman \cite{Simonyan2014} and Feichtenhofer et al. \cite{Feichtenhofer2016} both utilize optical flow combined with object detection in order to detect actions performed by humans. Giannakopoulos et al. also considered motion in their work for the purpose of detecting violent scenes in films \cite{Giannakopoulos}. Xu et al. used unsupervised learning to detect events partly based on motion, when analysing scenes of pedestrians walking \cite{Xu2015}.

Sports is a popular domain for event detection research. Ren et al. studied highlight detection in soccer games, studying 4 matches \cite{Ren2007} with good results, especially when detecting goals scored. Xu and Chua used not just audio-visual features but also external, text based, information in their work towards the detection of highlights in team sports \cite{Xu2006}. A similar approach, applied to baseball games, is proposed by Chiu et al. \cite{Chiu2012}. Sun et al. in  \cite{Sun2010} analysed the excitement level of sports commentators using audio features, mainly Mel Frequency Cepstrum Coefficients (MFCCs) and pitch data, to detect highlights.
Nguyen and Yoshitaka \cite{Nguyen2014} adopt a cinematography and motion based approach, whereby they analysed the type of camera shots used in order to detect highlights, especially emotional events. 

We use a measure of novelty to identify salient points in a stream. Novelty detection, including reconstruction error based systems, has been used in a wide range of other domains. Pimentel et al.'s review of novelty detection \cite{Pimentel2014} provides a comprehensive overview.

\subsection{Emotion Detection}
\looseness-1Studying streamers is, in part, the study of humans reacting to stimulus in an interactive setting. Therefore, whilst this work aims to develop event detection techniques, it is useful to consider work related to social and emotional signal processing, as it informs our approach.  Related work includes research in analyzing player experience during gameplay.  For example Karpouzis et al. developed the "Platformer Experience Dataset", that contains audio-visual material of subjects playing a platformer game, \textit{Infinite Mario Bros} \cite{7344647}.  The dataset has been utilized in several studies. For example, Shaker et al. \cite{Shaker2013}  develop player experience modeling techniques, while Asteriadis et al. \cite{Asteriadis2012} used this data set to develop techniques for clustering player types, with findings pointing to head movement being an indicator of player experience and skills.

Many affective computing techniques are related to those used in this  study, for example EmoNets \cite{Kahou2015} use Convolutional Neural Networks for understanding facial cues. Ghosh et al. use Fourier Coefficients and MFCCs fed to a variety of autoencoders for learning affect from speech \cite{Ghosh2015}. Similarly Amer et al. used extracted audio features, decomposed using Principal Component Analysis,  as input to a selection of deep networks \cite{6854297}. Busso et al. discuss frequency and emotion detection, confirming the result that pitch is an important indicator for emotion \cite{Busso2009}.

\subsection{Video Game Scene Analysis}
Little study has been undertaken into analyzing and extracting information from game scenes. The majority focuses on understanding the strategy, structure and physics of game worlds. For example, Guzdial and Riedl developed unsupervised techniques for building full game levels from observing gameplay videos \cite{Guzdial2016}. Croxton and Kortemeyer studied the way players learnt about physics related game content through the study of game play videos \cite{0031-9120-53-1-015012}. Lewis et al. used \textit{Starcraft} replays to discover strategies \cite{Lewis2011}. Alvernaz and Togelius used the latent space of an auto-encoder encoder to evolve agents for Visual Doom \cite{8080408}. 
Similarly Rioult et al. used the players in-game positions to predict winners in \textit{Defense of the Ancients} \cite{Rioult2014}.

\section{Methodology}
\subsection{Face and Game Scene Analysis}
In our work, we utilize convolutional autoencoders for analyzing both player face and game footage. The networks are composed of two stacked \text{VGG16}-like networks \cite{Simonyan2014_2}, omitting the fully-connected layers. Given a video frame, the encoder produces a $512$-filter encoding. The decoder is similar, employing reversed layers and up-sampling rather than max-pooling, reconstructing the input image.  Each convolutional layer has a $3\times3$ filter window and each max-pooling layer uses $2\times 2$ window with a stride of two, following works such as Deep Convolutional Auto-Encoder with Pooling - Unpooling layers, proposed by Turchenko et al. \cite{Turchenko2017}, and Stacked What-Where Auto-Encoders, proposed by Zhao et al. \cite{Zhao2015}. The network was trained using an \texttt{ADADELTA} optimiser \cite{Zeiler2012} using \texttt{Tensorflow} and \texttt{Keras}.  The reconstruction errors $R^{(1)}$ and $R^{(2)}$ (as shown in Fig. \ref{fig:sysoverview}) were utilized as indicators of novelty.  The complete network architecture is described in Table \ref{table:ae_arch}.

We note that for the face autoencoder, we used the \texttt{VGG} Face Descriptor weights \cite{Parkhi2015,Cao2017} for the encoder, which were frozen during training as no noticeable improvement was observed when fine-tuning and training end-to-end.  The autoencoder for game footage was trained end-to-end for each video. 

\begin{table}[t]
\resizebox{1\columnwidth}{!}{%
 \begin{tabular}{||c|c|c|| c ||c|c|c|| } 
 \cline{1-3}  \cline{5-7}
   Layer & Input & Output & & Layer & Input & Output \\
 \cline{1-3}  \cline{5-7}
 \cline{1-3}  \cline{5-7}
 Conv2d & (224,224,3) & (224,224,64) & & UpSample & (7,7,512) & (14,14,512) \\
 \cline{1-3}  \cline{5-7}
Conv2d & (224,224,64) & (224,224,64) & & Conv2d & (14,14,512) & (14,14,512) \\
\cline{1-3}  \cline{5-7}
MaxPool & (224,224,64) & (112,112,64) & & Conv2d & (14,14,512) & (14,14,512) \\
\cline{1-3}  \cline{5-7}
Conv2d & (112,112,64) & (112,112,128) & & Conv2d & (14,14,512) & (14,14,512) \\
\cline{1-3}  \cline{5-7}
Conv2d & (112,112,128) & (112,112,128) & & UpSample & (14,14,512) & (28,28,512) \\
\cline{1-3}  \cline{5-7}
 MaxPool & (112,112,128) & (56,56,128) & & Conv2d & (28,28,512) & (28,28,512) \\
\cline{1-3}  \cline{5-7}
Conv2d & (56,56,128) & (56,56,256) & & Conv2d & (28,28,512) & (28,28,512) \\ 
\cline{1-3}  \cline{5-7}
Conv2d & (56,56,256) & (56,56,256) & & Conv2d & (28,28,512) & (28,28,512) \\
\cline{1-3}  \cline{5-7}
Conv2d & (56,56,256) & (56,56,256) & & UpSample & (28,28,512) & (56,56,512) \\
\cline{1-3}  \cline{5-7}
MaxPool & (56,56,256) & (28,28,256) & & Conv2d & (56,56,512) & (56,56,256) \\
\cline{1-3}  \cline{5-7}
Conv2d & (28,28,256) & (28,28,512) & & Conv2d & (56,56,256) & (56,56,256) \\
\cline{1-3}  \cline{5-7}
Conv2d & (28,28,512) & (28,28,512)  & & Conv2d & (56,56,256) & (56,56,256) \\
\cline{1-3}  \cline{5-7}
Conv2d & (28,28,512) & (28,28,512) & & UpSample & (56,56,512) & (112,112,512) \\
\cline{1-3}  \cline{5-7}
MaxPool & (14,14,512) & (14,14,512) & & Conv2d & (112,112,256) & (112,112,128) \\
\cline{1-3}  \cline{5-7}
Conv2d & (14,14,512) & (14,14,512) & & Conv2d & (112,112,128) & (112,112,128) \\
\cline{1-3}  \cline{5-7}
Conv2d & (14,14,512) & (14,14,512) & & UpSample & (224,224,128) & (224,224,128) \\
\cline{1-3}  \cline{5-7}
Conv2d & (14,14,512) & (14,14,512) & & Conv2d & (224,224,128) & (224,224,64) \\
\cline{1-3}  \cline{5-7}
MaxPool & (14,14,512) & (7,7,512) & & Conv2d & (224,224,64) & (224,224,64) \\
\cline{1-3}  \cline{5-7}
\multicolumn{4}{ c ||}{ } & Conv2d & (224,224,64) & (224,224,3) \\ \cline{5-7}
 \end{tabular}%
 }
\caption{Auto-encoder layers. Left: encoder, right: decoder.}
\label{table:ae_arch}
\end{table}
Once trained on frames from a video, the reconstruction error can be used as an indicator for novel frames in a video - which in the context of this work, we consider as proxies for highlights.  More details on reconstruction-based novelty detection can be found in a recent survey by Primentel et al. \cite{Pimentel2014}.

\subsection{Audio Stream Analysis}
Since we are mostly interested in detecting arousal, we consider an approach that focuses on key audio frequencies.  In order to do so, we firstly consider $400$ms windows, with a sampling rate of $10$ samples per second, thus having a $300$ms overlap between sequential windows. A large enough window was utilized in order to alleviate issues arising from asynchrony between audio and video modalities, due to the preservatory and anticipatory co-articulation phenomena entailing that audio cues are often delayed by around $0.12$ seconds \cite{Katsaggelos2015}.

For each window, we compute the Short-Term Fourier Transform. We retain only the magnitudes (denoted as $\textbf{T}$ in Figure \ref{fig:sysoverview}), since they are deemed as good indicators for arousal \cite{doi:10.1111/1467-8721.00013}.  

Additionally, in order to isolate the streamers voice from the game sounds we discard frequencies which are uncommon in human speech\footnote{A more eloquent approach could be based on methods such as Independent Component Analysis (ICA)}. Frequencies between $300$ - $3400$ hertz were retained, a range chosen because it is the standard frequency range for telephones calls.  To capture variation while reducing dimensionality and noise, we apply Principal Component Analysis (PCA) and retain only the $k$ first principal components.

\subsection{Recurrent Layer for Late Fusion}
A recurrent layer is utilized for indicating reconstruction-based novelty across all views and modalities, which in-turn can be considered as a proxy for detecting stream highlights. The recurrent layer is fed with time-series consisting of reconstruction errors from face and game scene autoencoders, as well as the extracted audio features, in-effect performing a multi-dimensional smoothing operation.  Values are normalized between $[0,1]$ to avoid biasing views that cover different ranges, while training entails forecasting values corresponding to the next frame, $t+1$ at time $t$, thus effectively incorporating information from all views in a temporal late-fusion setting.

We utilize two Long Short-Term Memory (LSTM) layers \cite{lstm}, and a fully connected layer with a sigmoid activation. Each neuron in the LSTM layer retains a latent state capturing useful information from previous time frames, and along with the current input generates the output values. LSTMs also include a "forget gate" which allows the neuron to filter which input values are kept and which are discarded, improving training over a standard RNN layer. The network architecture is illustrated in Figure \ref{fig:sysoverview}.  We utilize the \texttt{ADADELTA} optimizer for training, employing a mean-squared error loss function.  The architecture is similar to the one utilized by Malhotra et al. \cite{Malhotra}.

\subsection{Highlight Detection as Reconstruction-based Novelty}
We utilize the prediction error of the recurrent layer  on a given stream session (including face, audio, and game footage) as an indicator of {\it stream highlights} ($E_i$ in Figure \ref{fig:sysoverview}).  In more detail, we utilize a threshold, empirically determined as $0.01\%$, and classify the same percentage of frames with highest prediction error as {\it highlight frames}. We use the aforementioned highlight frames in order to generate the full highlight clips.  To do so, we treat the detected frames as {\it apex frames} of a highlight event.  We link proximal apex frames together\footnote{Apex frames are linked if not doing so would cause an overlap between clips.}, and similarly to the \texttt{TWITCH.TV} clip system, consider the highlight clip to be $10$ seconds before the apex frame and $5$ seconds after the last.  In this way, we ensure that the appropriate context is included in the highlight clip, and that the clip is self-contained (e.g., a reaction of the streamer can be detected as a highlight frame, with a preceding game or stream event causing the reaction).

\section{Data}
Data was gathered from \texttt{TWITCH.TV}. We recorded popular streamers playing \textit{Player Unknown's Battlegrounds} (PUBG), a multi-player on-line battle royale game. A number of players are spawned simultaneously with a goal of exploring an island, collecting weapons, killing other players and ultimately being the last player alive. PUBG was chosen because it often has long periods of low intensity game-play and short bursts of concentrated action, making highlight detecting a worthwhile task, while an abundance of high-quality streams are available due to the popularity of the game.

Each recorded stream was segmented into videos spanning a single game, and downtime between games removed. It makes little sense to look for highlights when the game is not being played as streamers often take short breaks in-between games where they will leave the stream or browse social media etc. The data set consists of videos from two streamers, both male, one American and one German but streaming in English. There is a total of $11$ videos and each video is between $19$ minutes $30$ seconds and $30$ minutes $40$ seconds long.  In total, we utilize over $5$ hours of stream footage. 

We pre-process each video as follows.  Firstly, we utilize a sampling rate of $10$ frames per second, which is deemed sufficient for our task and makes training faster.  We mask-out the players face for feeding the game footage into the respective autoencoder, while the cropped region including the players face is used for the face autoencoder.  Finally, we resize each frame to $224\times 224 \times 3$ in order to match the \texttt{VGG} Face Descriptor dimensions. 
\section{Results}
In total, we obtained $98$ segmented highlight clips by applying our method on $11$ game stream recordings.  To evaluate our method, we manually annotated each highlight clip into $4$ categories, namely ``funny'', ``action'', ``social-interaction'', and finally, ``no highlight''.  Funny videos are streamer-focused events where the streamer makes a joke, laughs, or is in other ways amused.  Action highlights stem from game-events (e.g., streamer engaged in a firefight).  Highlights that are tagged as ``social-interaction'' include events that are community-led, where the streamer interacts with viewers in a meaningful way, e.g., thanking subscribers, answering questions, or reacting during a similar interaction.  Note that ``social-interaction'' highlights are important for a compelling game stream, and are often found in streamer highlight clips that are manually segmented.  Finally, clips containing no noteworthy events are labeled as ``No Highlight''.   In Table \ref{table:1}, we show results by using all available modalities, where out of $75$ clips with interesting content, $51$ are tagged as``funny'' or ``action'', with the remaining $24$ labeled as ``social-interaction''.

\begin{table}[t]
\begin{center}
\resizebox{0.8\columnwidth}{!}{%
 \begin{tabular}{||c| c| c |c|c|c||} 
 \hline 
  \multirow{2}{*}{Video} & \multicolumn{4}{| c |}{Highlight}  & No Highlight  \\ 
 \cline{2-6}
   & Funny & Action & Interaction & Total & Total \\
 \hline\hline
 S1\_1 & 1 & 2 & 4 & 7 & 0 \\ 
 \hline
 S1\_2 & 0 & 1 & 1 & 2 & 4 \\
 \hline
 S1\_3 & 0 & 3 & 2 & 5 & 3 \\
 \hline
 S1\_4 & 2 & 2 & 4 & 8 & 2 \\
 \hline
 S1\_5 & 0 & 3 & 4 & 7 & 4 \\ 
 \hline
 S1\_6 & 0 & 1 & 1 & 2 & 2 \\ 
 \hline
 \hline
 S2\_1 & 5 & 2 & 0 & 7 & 1 \\ 
 \hline
  S2\_2 & 3 & 1 & 2 & 6 & 1 \\ 
 \hline
  S2\_3 & 6 & 4 & 2 & 12 & 2 \\ 
 \hline
  S2\_4 & 3 & 1 & 3 & 7 & 1 \\ 
 \hline
  S2\_5 & 6 & 5 & 1 & 12 & 3 \\ 
 \hline
 \hline
   Total & 26 & 25 & 24 & 75 & 23 \\ 
   \hline
\end{tabular}}
\end{center}
\caption{Generated highlight clips by category using all modalities and views.}
\label{table:1}
\end{table}

\begin{figure*}[t]
\begin{center}
\includegraphics[width=1\textwidth]{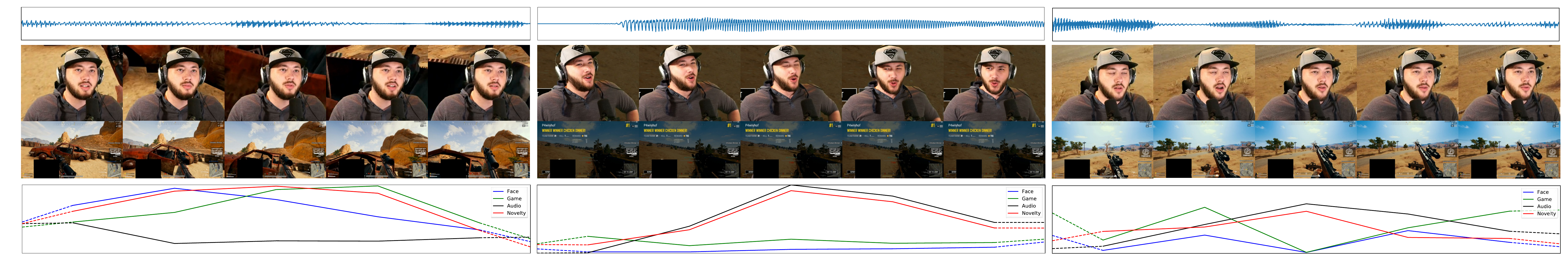}
\caption{Example highlights discovered by the proposed method. Left: The streamer begins aiming their rifle which changes the game scene enough to trigger a highlight, in agreement with a change in the facial expression. Center: The streamer wins a game and shouts in celebration; detected by indicating novelty in the audio features. Right: During a firefight, gunfire causes a spike in the audio which triggers a highlight.}
\label{fig:results}
\end{center}
\end{figure*}

\subsection{Modalities and Highlight Detection}
We evaluate the proposed architecture when observing different combinations of views and modalities, including ``Face'', ``Game Footage'', and ``Audio'', with results summarized In Table \ref{table:summary}.  Overall, we find that the model fusing all views and modalities performs the best.  This is an expected result, since utilizing audio-visual information from both streamer behaviour and the game itself is deemed to provide a more informed approach.  We also note  that the number of detected highlights across all combinations is similar, with the exception of the ``Audio Only'' model, that produced considerably more.  This is likely due to the impact of in-game audio (e.g., gunfire) that has not been entirely removed - further supported by the observation that  $29\%$ of highlights selected were action highlights, containing only a few funny or interaction clips.  The ``Face Only'' stream is better at determining ``funny'' and ``social-interaction'' highlights, although has a worse precision in terms of action clips, which is expected given that using only the face means we do not consider any context regarding the game. ``No Highlight'' clips are comparable for the audio-only and face-only models, and are often due to unusual gestures that the streamer might perform (that can potentially occlude the face). This is expected since ``face novelty'' itself does not necessarily indicate a highlight.  

The game-footage model is the worst highlight predictor, given our results.  Whilst this might appear counter-intuitive, we can also see that using game footage correctly detects the vast majority of action highlights\footnote{Although it is possible that on-screen events are indicators of other highlight types, for example new subscriber pop-ups or humorous on-screen events.}.  The lack of social- and game- context or scene understanding makes the problem of detecting when the game-scene is interesting to the viewer rather than merely anomalous more challenging.

Finally, we present results using a face and audio model.  This approach provides better results than all single-view models, and slightly worse results compared to using all available views and modalities.  This finding suggests that whilst the game alone is a poor indicated of highlights (other than action clips), it can be useful to corroborate information extracted from other modalities, and improve performance.

\subsection{Highlights over Time}
\begin{figure}[t]
\includegraphics[width=1\columnwidth]{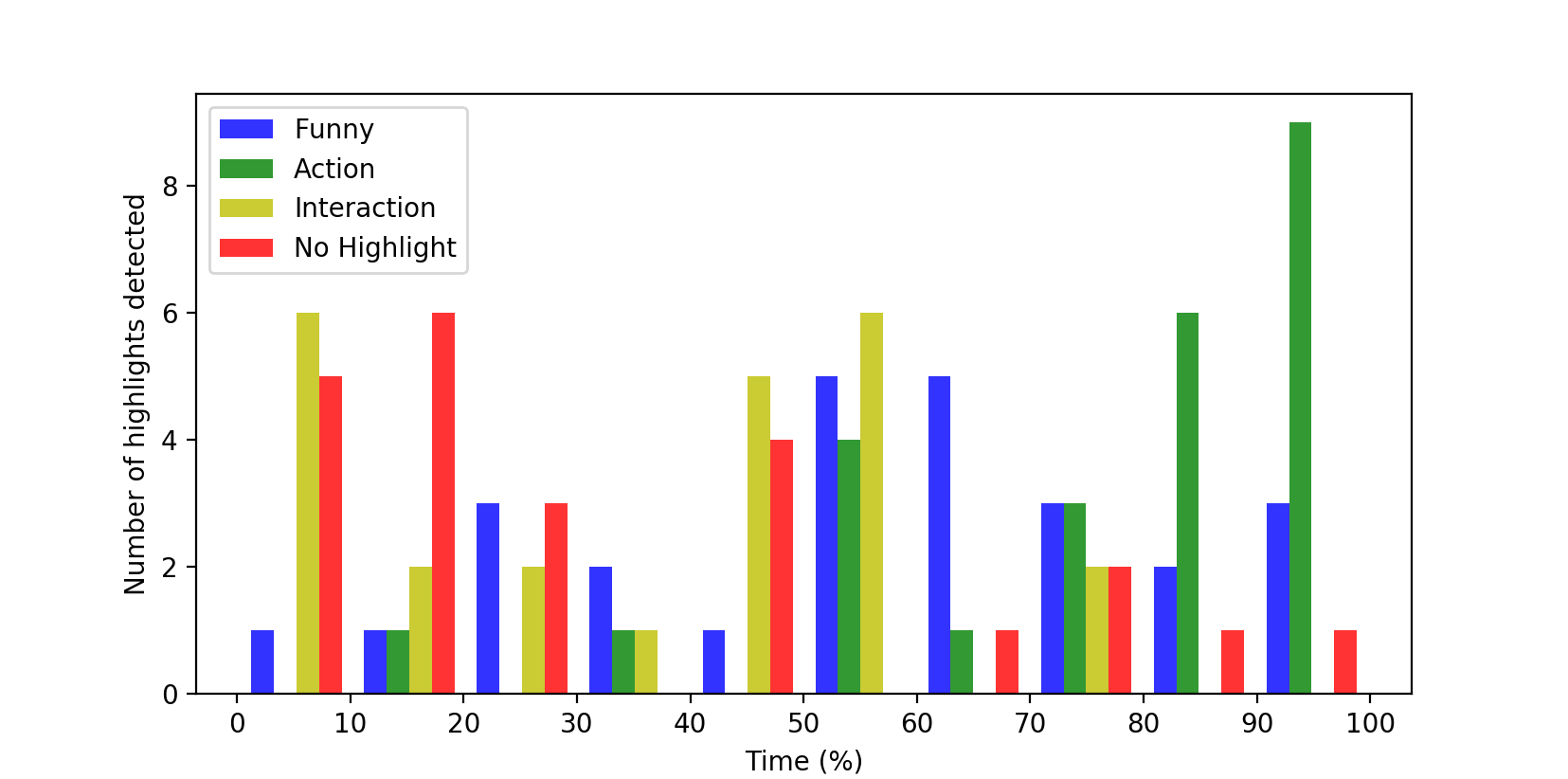}
\caption{Highlights by type over time}
\label{fig:highlights_overtime}
\end{figure}

We observe that the number of ``No Highlight'' segments is reduced over time, as shown in Figure \ref{fig:highlights_overtime}.  This points to the conclusion that the later in the video a highlight is detected, the more likely it is to be interesting. In more detail, $61\%$ of "No Highlight" results occur in the first $30\%$ of a video, opposed to $19\%$ of funny clips, $4\%$ off action clips and $41\%$ of interaction clips.  Furthermore, the majority of action clips, $92\%$, and funny clips, $69\%$, occur in the last $50\%$ of the video, opposed to only $22\%$ of "No Highlights" clips. In fact $60\%$ of action clips occur in the last $20\%$ of the video duration. By considering only detections in the last half of each video, we find that $91\%$ of clips are interesting in some way. 

Based on our observations of the streams, we can attribute this to several reasons that are mostly related with game design.  Firstly, the game is designed in such way the the play area shrinks over time, in a way that forces interaction between players towards the end of a game, hence the larger amount of action highlights towards the end of the video.  Secondly, there are fewer viewer interactions as the game progresses, since the game intensity increases and players are required to focus more on the game.

\begin{table}[t]
\begin{center}
\resizebox{1\columnwidth}{!}{%
 \begin{tabular}{||c| c |c| c |c|c|c||} 
 \hline 
  \multirow{2}{*}{Modalities} &  \multirow{2}{*}{No. Videos} & \multicolumn{4}{| c |}{Highlight}  & No Highlight  \\ 
 \cline{3-7}
   & & Funny \% & Action\% & Interaction\% & Total\% & Total\% \\
 \hline\hline
 Face, Game, Audio & 98 &0.27 & 0.26 & 0.24 & 0.77 & 0.23 \\ 
 \hline
 Face, Audio &95 &0.22 & 0.23 & 0.28 & 0.74 & 0.26\\
 \hline
 Face Only & 96&0.14 & 0.14 & 0.24 & 0.52 & 0.48 \\
 \hline
 Game Only &94& 0.04 & 0.18 & 0.07 & 0.29 & 0.70\\
 \hline
 Audio Only &126& 0.08 & 0.29 & 0.18 & 0.56 & 0.44\\
 \hline
\end{tabular}}
\end{center}
\caption{Summary comparison of highlight-detection over multiple views and modalities}
\label{table:summary}
\end{table}

\subsection{Novelty Across Modalities}
\begin{figure}[t]
\begin{center}
 \begin{subfigure}[b]{0.49\columnwidth}
\includegraphics[width=\textwidth]{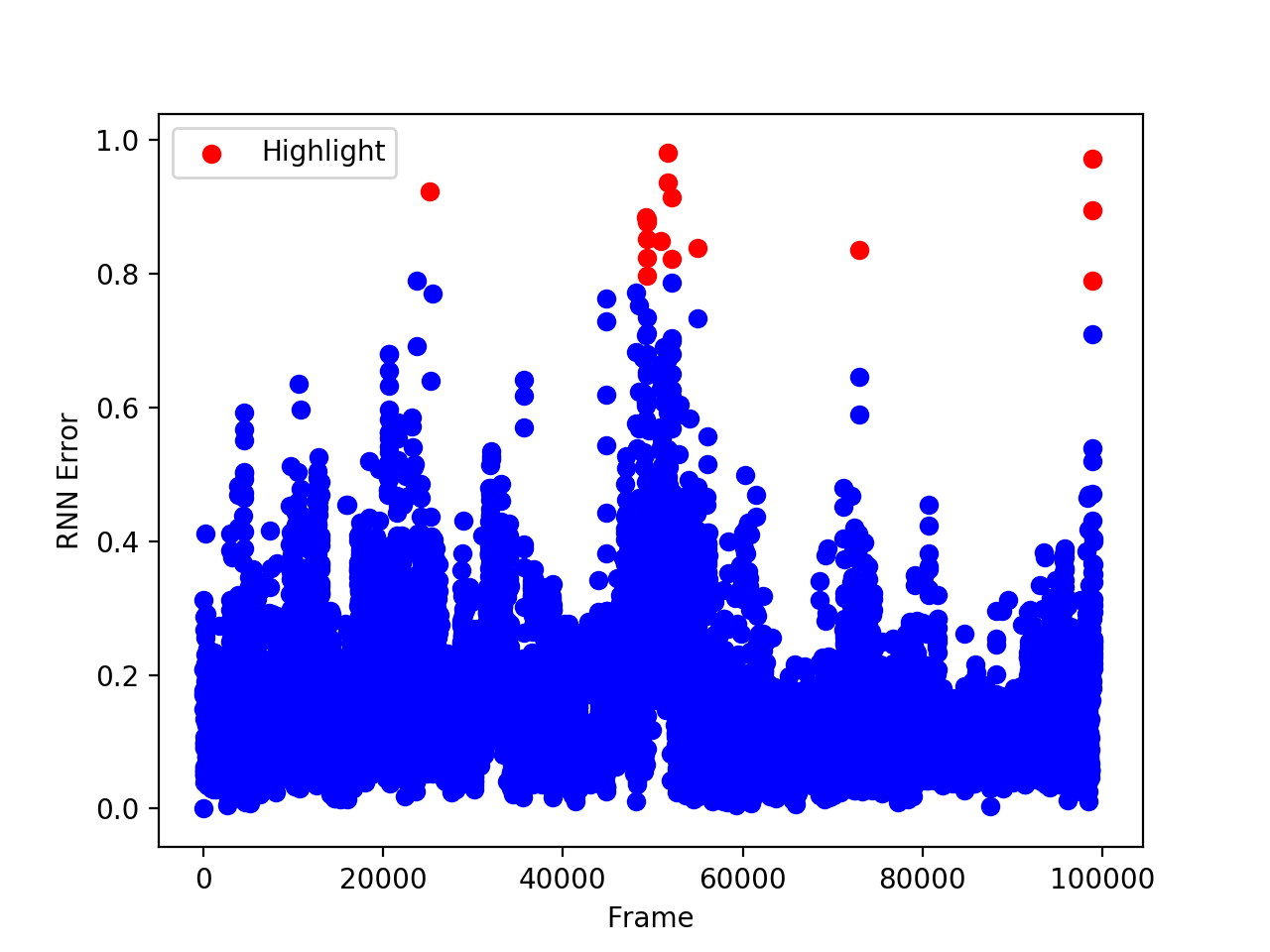}
\caption{Fused}
 \end{subfigure}
  \begin{subfigure}[b]{0.49\columnwidth}
 \includegraphics[width=\textwidth]{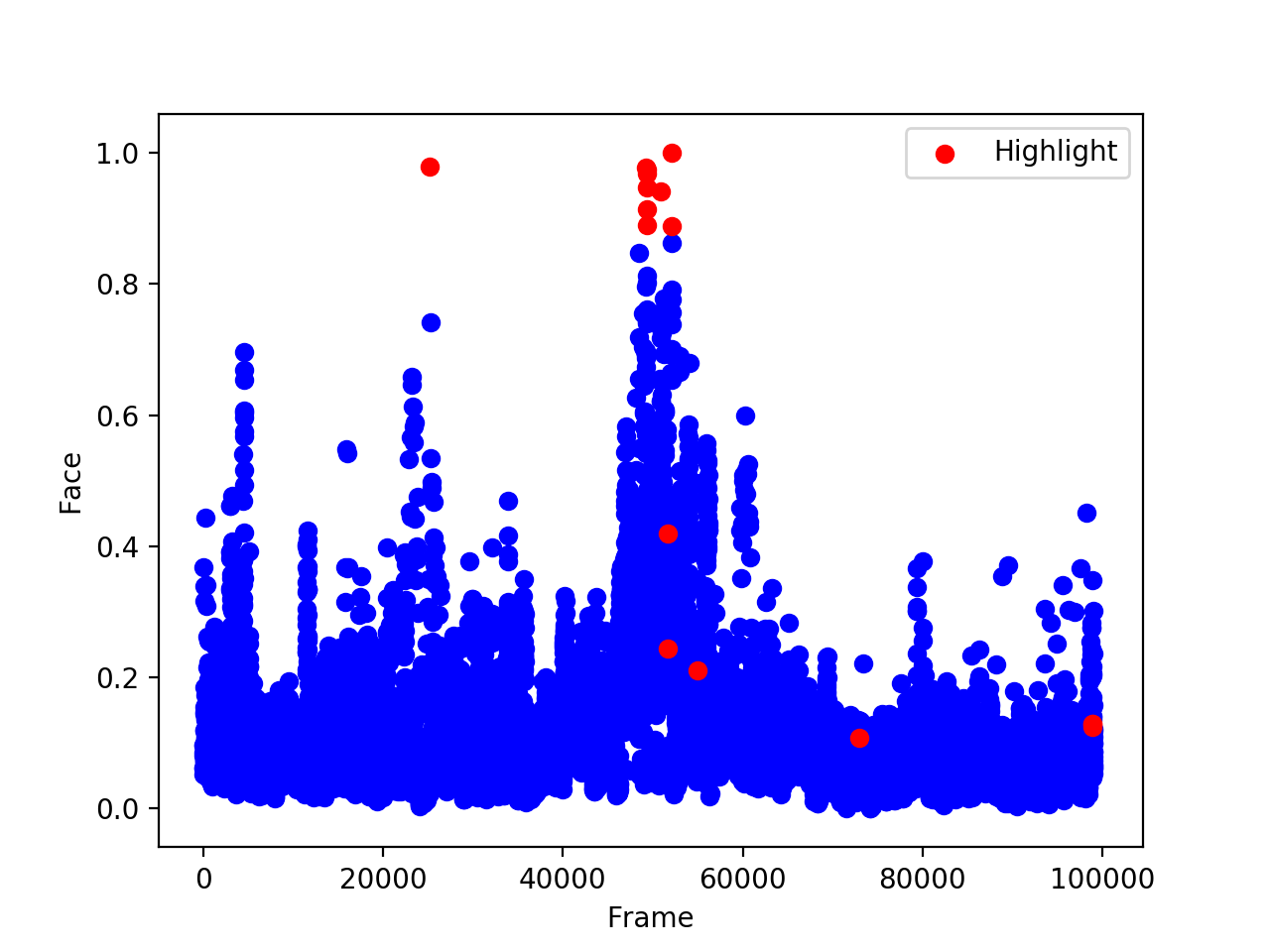}
 \caption{Face}
 \end{subfigure}
  \begin{subfigure}[b]{0.49\columnwidth}
 \includegraphics[width=\textwidth]{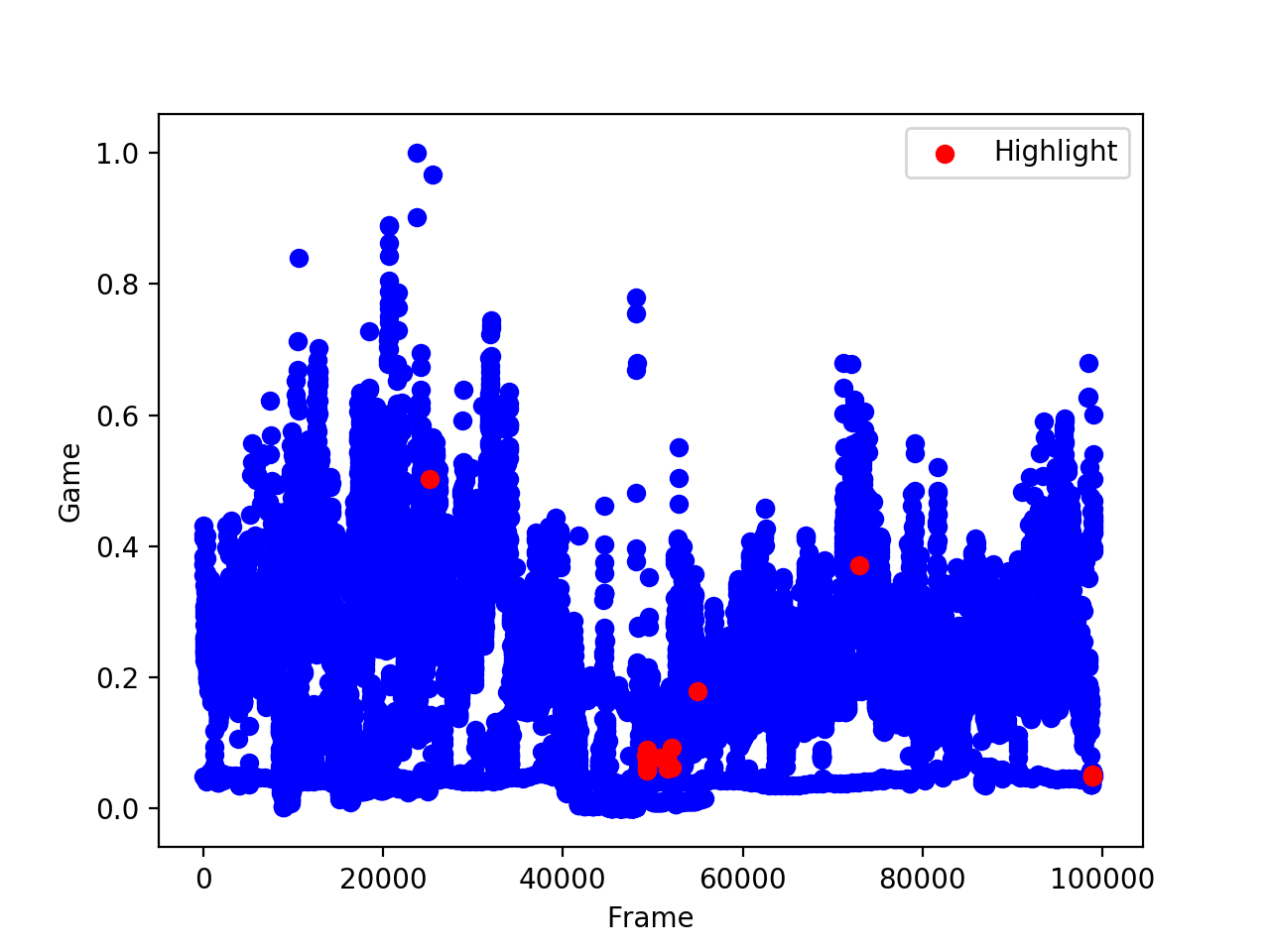}
 \caption{Game}
 \end{subfigure}
  \begin{subfigure}[b]{0.49\columnwidth}
\includegraphics[width=\textwidth]{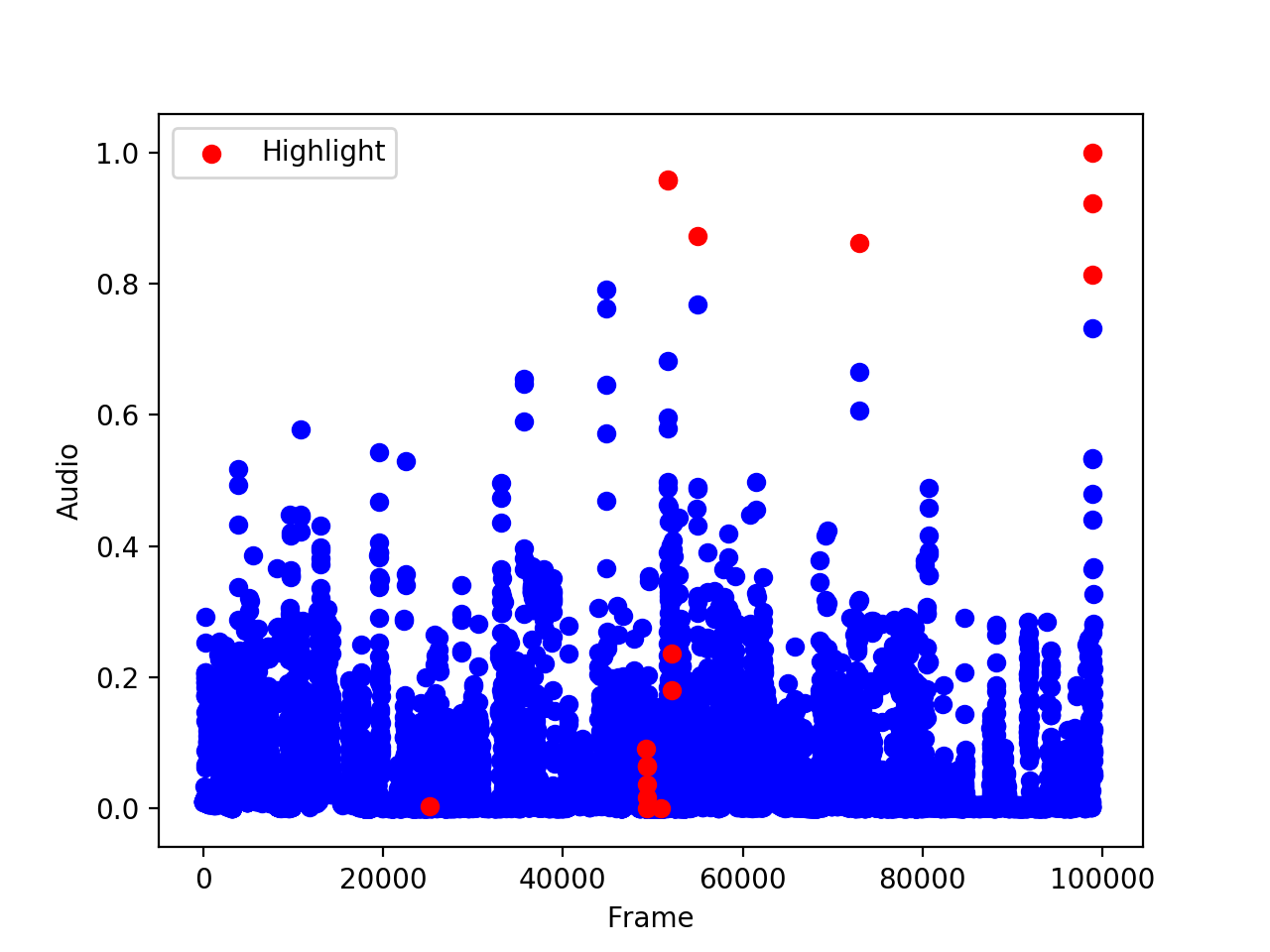}
\caption{Audio}
 \end{subfigure}

\caption{Errors over time indicating novel events for a particular video (\texttt{S1\_1}).  (a) Fused prediction error. (b) Face video reconstruction error. (c) Game footage reconstruction error. (d) Audio features over time.}

\label{fig:input_output}
\end{center}
\end{figure}

In this section, we discuss the detection of novel events across modalities.  In Figure \ref{fig:input_output}, we show (a) the RNN prediction error fusing all modalities and views, (b,c) the face and game footage autoencoder reconstruction errors, and (d) the first principal component of the Fourier coefficients of the audio channel.  We plot the errors over time for a particular video \texttt{S1\_1}, while coloring errors that correspond to selected highlight frames in red.  In general, we can observe that for face and game, sharp ``spikes'' pointing to highlights can be clearly observed in the distribution, with errors on game footage being less clear, showing a higher average and wider spread, likely due to the lack of a baseline/context.  For this particular stream, we can also observe that the game footage impacts the final highlight frames {\it less} than the face and audio views, where clear spikes are transferred to the fused results.  Observing the final results, we can see that spikes in only one view are smoothed out, while spikes appearing in more than one views are accentuated.

\section{Conclusions}
We presented an unsupervised deep learning architecture for  detection of highlight clips based on audio-visual data, broadcasted during a typical game stream. We consider a measure of reconstruction-based novelty as a proxy for indicating highlights, while {\it jointly} analyzing facial footage of the player, footage of the streamed game, as well as audio.  We discuss several insights arising from our analysis, while we show that the proposed method is successful in terms of detecting both social and game-related highlights in video game streams, further pinpointing the significance of considering social signals towards detecting interesting highlights in game streams. Future works into this domain would include widening the study to a more streamers playing a wide range of games. 

\appendix
\begin{acks}
This work was supported by EPSRC grant EP/L015846/1 (IGGI). We thank the NVIDIA Corporation for providing a Titan X Pascal GPU used in this work.  The authors would also like to thanks Mats Kathage (P4wnyhof) for permission to present footage from his stream in this work.
\end{acks}

\bibliographystyle{ACM-Reference-Format}
\bibliography{bibliography}
\end{document}